\documentclass[a4paper, 10pt, conference]{ieeeconf}      

\usepackage{FG2024}
\FGfinalcopy 

\IEEEoverridecommandlockouts                              
\usepackage{graphics} 
\usepackage{graphicx}
\usepackage{diagbox}

\usepackage{amsmath} 
\usepackage{multirow}
\usepackage{rotating}

\usepackage{hyperref}

\def\FGPaperID{126} 

\title{\LARGE \bf
$\epsilon$-Mesh Attack: A Surface-based Adversarial Point Cloud Attack for Facial Expression Recognition 
}

\author{\parbox{16cm}{\centering
    {\large Batuhan Cengiz$^1$, Mert Gülşen$^1$, Yusuf H. Sahin$^2$, Gozde Unal$^1$}\\
    {\normalsize
    $^1$ Department of AI and Data Engineering, Istanbul Technical University, Istanbul, Turkey\\
    $^2$ Department of Computer Engineering, Istanbul Technical University, Istanbul, Turkey}}
    \thanks{This work was supported by the National Center for High Performance Computing of Turkey (UHEM), under the Grant Number 1007422020.}
}

\usepackage{fancyhdr}
\begin{document}

\ifFGfinal
\thispagestyle{empty}
\pagestyle{empty}
\else
\author{Anonymous FG2024 submission\\ Paper ID \FGPaperID \\}
\pagestyle{plain}
\fi
\maketitle

 \thispagestyle{fancy}

\begin{abstract}

Point clouds and meshes are widely used 3D data structures for many computer vision applications. While the meshes represent the surfaces of an object, point cloud represents sampled points from the surface which is also the output of modern sensors such as LiDAR and RGB-D cameras. Due to the wide application area of point clouds and the recent advancements in deep neural networks, studies focusing on robust classification of the 3D point cloud data emerged. To evaluate the robustness of deep classifier networks, a common method is to use adversarial attacks where the gradient direction is followed to change the input slightly. The previous studies on adversarial attacks are generally evaluated on point clouds of daily objects. However, considering 3D faces, these adversarial attacks tend to affect the person's facial structure more than the desired amount and cause malformation. Specifically for facial expressions, even a small adversarial attack can have a significant effect on the face structure. In this paper, we suggest an adversarial attack called $\epsilon$-Mesh Attack, which operates on point cloud data via limiting perturbations to be on the mesh surface. We also parameterize our attack by $\epsilon$ to scale the perturbation mesh. Our surface-based attack has tighter perturbation bounds compared to $L_2$ and $L_\infty$ norm bounded attacks that operate on unit-ball. Even though our method has additional constraints, our experiments on CoMA, Bosphorus and FaceWarehouse datasets show that $\epsilon$-Mesh Attack (Perpendicular) successfully confuses trained DGCNN and PointNet models $99.72\%$ and $97.06\%$ of the time, with indistinguishable facial deformations. 
The code is available at \href{https://github.com/batuceng/e-mesh-attack}{\color{magenta}https://github.com/batuceng/e-mesh-attack}.

\end{abstract}

\section{INTRODUCTION} \label{INTRODUCTION}

Advances in deep learning  eased many tasks defined on 3D point clouds and mesh structures such as classification \cite{qi2017pointnet,qi2017pointnet++,deng2018ppfnet, wang2019dynamic}, generation \cite{kim2021setvae,kim2020softflow,mo2023dit} and registration \cite{hezroni2021deepbbs, huang2020feature, wang2019deep}. However, solving such complicated problems with deep models has raised robustness concerns \cite{carlini2017towards} and adversarial attacks \cite{goodfellow2014explaining} are suggested to evaluate robustness. Adversarial attacks aim to generate data examples with imperceptible yet effective small perturbations to mislead vision models. 

There have been many studies on designing adversarial attacks for 2D \cite{madry2017towards, kurakin2018adversarial} and 3D \cite{xiang2019generating, yang2019adversarial, huang2022shape, zhang20233d} data. The crucial difference between 2D and 3D attacks is that 3D attacks perturb the point positions while 2D attacks change the pixel values, keeping the same positions. By performing an attack on a point cloud, a slightly jittered version of the original point cloud is constituted which fallaciously makes the network predict the wrong class. While yielding great results in terms of accuracy by confusing deep learning models, existing methods fall short on preserving the surface structure, especially in 3D facial expression data since small deformations can cause the overall expression to change.

Preserving the surface structure can be crucial since capturing a facial expression via a 3D sensor could be done by sampling points over the face surface that is represented in the mesh. Thus, for many applications, the point cloud and the related mesh are available together \cite{colombo2011umb, yang2020facescape, cao2013facewarehouse, TMPEH:CVPR:2023, COMA:ECCV18}. However, previous 3D attack methods do not consider mesh data. Following these ideas, we are motivated to develop an adversarial attack method that preserves the face surfaces on 3D point clouds by utilizing available mesh data.

\begin{figure}
    \centering
    \includegraphics[width = 0.9\linewidth]{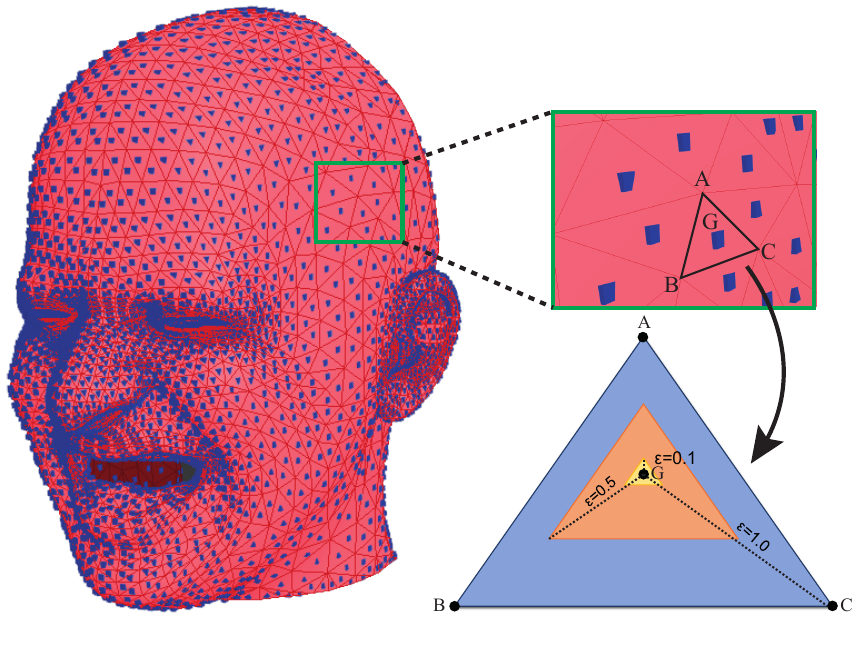}
    \caption{An example Face mesh from CoMA\cite{COMA:ECCV18} dataset and the suggested triangular bounds for the surface preserving white box attack scaled by parameter $\epsilon$.}
    \label{fig:example-coma-face}
\end{figure}

In this paper, we propose a 3D adversarial attack method for point clouds called \textit{$\epsilon$-Mesh attack}, which preserves the face surface by strictly keeping adversarial points on the mesh by projecting perturbations onto mesh triangles using two different methods: central and perpendicular projections. We have also parameterized our attack method by $\epsilon$ to scale our attack boundaries into similar triangles as shown in Fig.~\ref{fig:example-coma-face}. We evaluate our attack method on 3D facial expression recognition models and show that compared to other attacks, our attack does not cause as much surface deformation. This creates potential use cases for our method in many cases such as safety-critical application of facial expression recognition like human computer interaction or classification in the wild with unsafe data.



\section{RELATED WORK} \label{RELATED WORK}



\textbf{Facial Expression Recognition}. In the classical facial expression recognition problem, following Ekman's \cite{ekman1999basic} studies, initial methods investigated 2D images or image sequences to map the inputs to basic emotional expressions like anger, disgust and fear \cite{dhall2015video, lucey2010extended}.  In \cite{richter2012facial}, DCT, LBP, and Gabor filters are used to extract facial features, then an SVM is trained to classify the expressions. In \cite{afshar2016facial}, a neural network is trained with facial landmark trajectories and geometric features. In \cite{liu2017adaptive}, a deep metric learning-based training procedure is presented. In \cite{yang2018facial}, a generative network that outputs faces with no expression from expressive face images is trained, and then using the features from this network, a classification is done.

With the emergence of deep learning-based models that directly process point cloud data \cite{qi2017pointnet, qi2017pointnet++, wang2019dynamic}, there have been significant developments in 3D facial expression recognition. These permutation invariant point cloud-based models can learn to represent the structure of point cloud face data. In \cite{liu20234d}, geometrical images for 3D face sequences are fed to Dynamic Geometrical Image Network (DGIN) which combines short-term and long-term information. In \cite{duh2016facial}, histograms of oriented gradients and optical flows are utilized to find correspondence in 3D data and classify facial expressions. In \cite{behzad2021disentangling}, a multi-view transformer architecture is proposed for 3D/4D facial expression recognition. For these tasks, a variety of 3D datasets are accessible, including BU-3DFE \cite{yin20063d}, FaceScape \cite{yang2020facescape}, along with 4D datasets like BU4DFE \cite{yin20084d} and BP4D+ \cite{zhang2014bp4d}. For a further reading on 3D facial expression recognition, detailed surveys on this topic can be investigated \cite{liu20234d, li2020deep}.


\textbf{Adversarial Attacks}. Adversarial attacks are methods that generate adversarial data perceptually similar to the original samples, to deceive deep learning models. Szegedy \textit{et al.} \cite{szegedy2013intriguing} pioneered to demonstrate the vulnerability of neural networks to adversarial examples and drew attention to the potential security risks in safety-critical applications.
In the 2D image domain, many attack methods have been proposed for deep learning models. Goodfellow \textit{et al.} \cite{goodfellow2014explaining} argued that generation of adversarial examples are possible due to locally linear nature of neural networks and proposed an attack method referred to as Fast Gradient Sign method. This method allows generation of adversarial examples with one step towards the direction of gradient to increase the loss. Madry \textit{et al.} \cite{madry2017towards} demonstrated Projected Gradient Descent (PGD) attack by applying multiple gradient steps in a bounded area to find local minimum. Another attack algorithm called C\&W attack was proposed by Carlini \textit{et al.} \cite{carlini2017towards} which optimizes an objective function of distance between original and adversarial examples subject to the constraint of changing the classification of image in order to find the adversarial perturbation.

After their success on 2D images, adversarial attacks are extended to 3D point cloud models.
Xiang \textit{et al.} \cite{xiang2019generating} pioneered the extension of adversarial attack methods to 3D point cloud models by generating 3D adversarial point cloud examples using their proposed methods, adversarial point addition and perturbation.
Yang \textit{et al.} \cite{yang2019adversarial} proposed pointwise gradient perturbation, point attachment and detachment methods by leveraging gradient-based adversarial attack algorithms.
Zhang \textit{et al.} \cite{zhang20233d} proposed an attack method that minimizes combined loss of mesh edge distances and sampled point cloud Chamfer distances to directly create adversarial meshes. Huang \textit{et al.}\cite{huang2022shape} suggested differentiable rotation and translation matrices to create adversarial perturbations that lie on estimated surfaces. 
Projected Gradient Descent (PGD) attack in 3D \cite{sun2021adversarially} iteratively move the points towards gradient directions to maximize loss where the total translation is limited to a spherical $\epsilon$-ball in $L_2$ and $L_{\infty}$ distance metrics. Inspired by PGD attack, we have proposed two different projection mechanisms to limit the adversarial perturbations on the 2D mesh surfaces rather than 3D $\epsilon$-ball.

\addtolength{\textheight}{-3cm}   

\section{METHOD} \label{METHOD}


\textbf{Meshes \& Point Clouds}. A mesh $M$ is as a set of $v$-gons in the $d$-dimensional space. In standard mesh processing, the mesh is defined in the space of triangles ($v=3$) in 3D space ($d=3$). Thus, a mesh could be defined as $M = \{t_1,...,t_i,..., t_n\}$ where $t_i$ represent the triangles. From each triangle, a sampling process could be performed to create the point cloud $P = \{p_1,...,p_i,..., p_n\}$ where $p_i$ represents the point sampled from $t_i$. We should also note that this sampling process is independent for each triangle.

Moreover, We can define each triangle as $t_i = \triangle{A_iB_iC_i}$ where vertices are $A_i,B_i,C_i \in R^3$ and the barycenter is $G_i = (A_i+B_i+C_i)/3 \in R^3$. We also denote the normal vector as $\vec{n_i} \perp \triangle{A_iB_iC_i}$. 

Assuming that, each triangle has a uniform probability density function for the sampled point $p_i$, $E[p_i] = G_i$ is the most representative point of $t_i$ and could be used in sampling. We also followed this method to initialize our point clouds from the meshes.


\textbf{Adversarial Perturbations}. Given a classification function $f_\theta(P)$, its prediction $\hat{K}$, and the ground truth label $K$ we define the adversarial perturbation $\vec\nabla \in R^3$ as the gradient ascent step with respect to loss $L = \|K - \hat{K}\|$ as,

\begin{equation} \label{project nabla}
    \vec\nabla = \frac{dL}{df_\theta} \frac{df_\theta}{dP}. 
\end{equation}

\textbf{Projection Methods}. To keep the adversarial perturbation of $p_i$ on the triangle $\triangle{A_iB_iC_i}$, we have to project $\vec\nabla$ back into the plane using the formula:

\begin{equation} \label{project nabla}
    \vec\nabla_s = \vec\nabla - \frac{\vec\nabla \cdot \vec{n_i}}{\vec{n_i} \cdot \vec{n_i}} \vec{n_i} 
\end{equation}
where $\hat{p}_i = p_i + \alpha \vec\nabla_s$ is the projected position of point $p_i$ and $\alpha$ is the attack learning rate. It is possible for $\hat{p}_i$ to lie outside of the region bounded by triangle $\triangle{A_iB_iC_i}$. Thus, we propose the following two methods to project $\hat{p}_i$ back into the boundary of $\triangle{A_iB_iC_i}$ if it is not already in the triangle.

\begin{figure}[]
    \centering
    \includegraphics[width=0.5\linewidth]{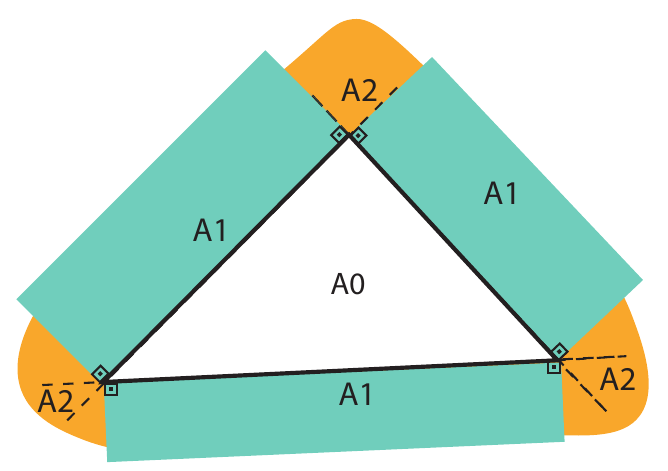}
    \caption{Division of the area near a selected triangle to calculate perpendicular projection.}
    \label{fig:ucgen-7}
\end{figure}

\textbf{Central Projection}. Firstly, we suggest projecting point $\hat{p}_i$ that is outside of the triangular region $\triangle{A_iB_iC_i}$ by the line that directs into barycenter $G_i$. Thus, the projection $p_{i, cent}$ is defined as the intersection point of line segment $\overline{\rm G_i\hat{p}_i}$ and intersecting edge of the triangle $\triangle{A_iB_iC_i}$ as given in Equation (\ref{eq_central}).

\begin{equation}
    p_{i,cent}= 
\begin{cases}
    \hat{p}_{i},&  \hat{p}_i \in \triangle{A_iB_iC_i}\\ \\
    \triangle{A_iB_iC_i} \cap \overline{\rm G_i \hat{p}_i},              & \text{otherwise}
\end{cases}
\label{eq_central}
\end{equation}

\textbf{Perpendicular Projection}. Secondly, we suggest the perpendicular projection method where the area near the triangle is divided into 7 parts as given in Figure \ref{fig:ucgen-7}. If $\hat{p}_i$ is in the $A0$ region, it is not projected. If $\hat{p}_i$ is inside an $A1$ region, it is projected to the nearest edge of the triangle. For the $A2$ regions, the nearest vertex of the triangle is selected. To formally define, if the projected point is outside of the triangle, it is projected to the closest point of the triangular area as shown in Equation (\ref{eq_perp}).

\begin{equation}
    p_{i,perp}= 
\begin{cases}
    \hat{p}_{i},&  \hat{p}_i \in \triangle{A_iB_iC_i}\\ \\
    \underset{x \in \triangle{A_iB_iC_i}}{\arg\min} \|x-\hat{p}_i\|,              & \text{otherwise}
\end{cases}
\label{eq_perp}
\end{equation}

\begin{figure}[t]
    \centering
    \includegraphics[width = 0.9\linewidth]{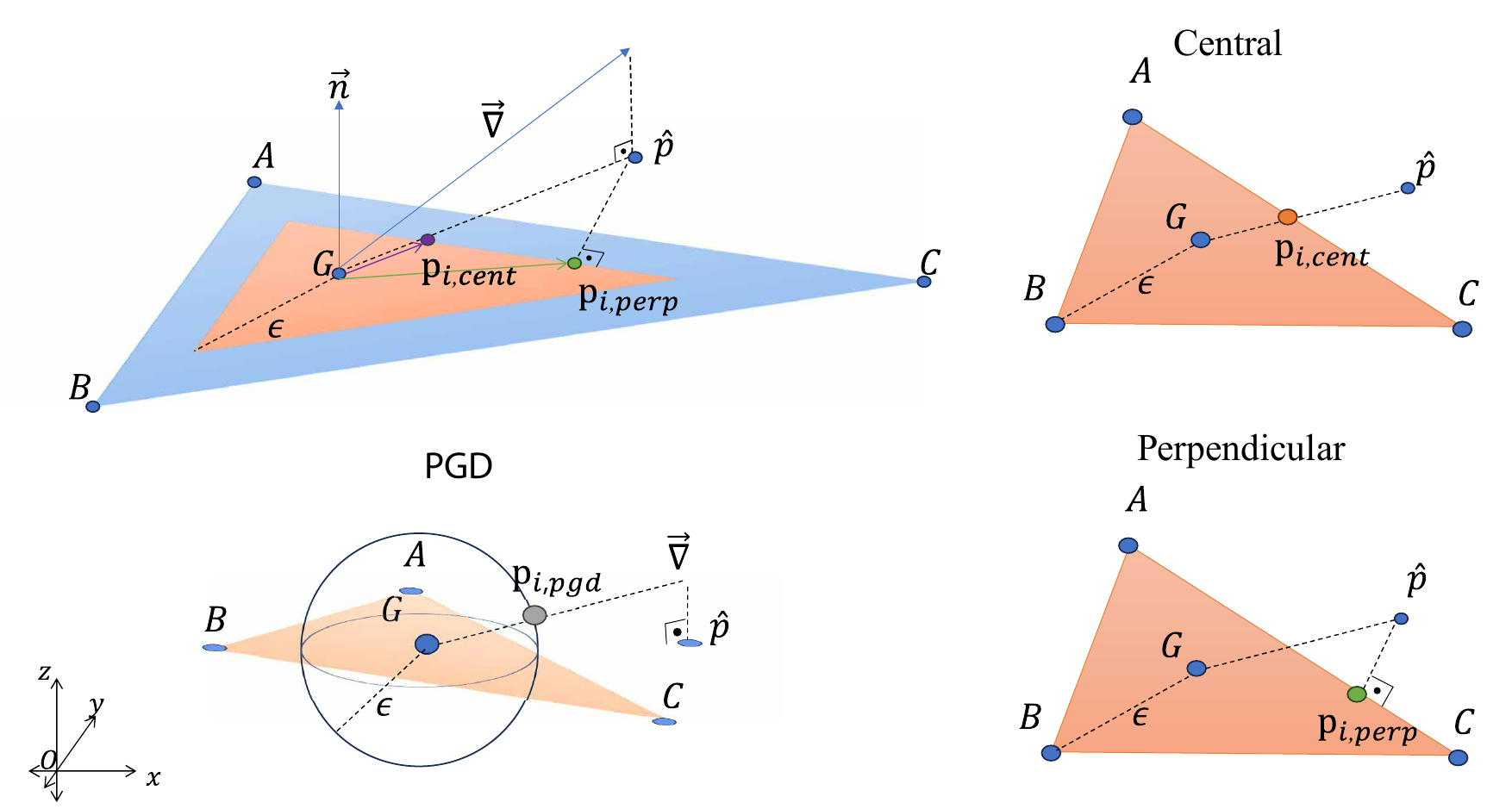}
    \caption{Projection example for adversarial perturbation $\nabla$ (left). Different projection methods in right: PGD\cite{sun2021adversarially}, Central and Perpendicular from top to bottom.}
    \label{fig:enter-label}
\end{figure}

A comparative demonstration of PGD and $\epsilon$-Mesh projections is given in Fig.~\ref{fig:enter-label}. The proposed projection methods are applied on each step of gradient ascent optimization process. The triangle $\triangle{A_iB_iC_i}$ can be scaled by an epsilon paramater $\epsilon \in [0, 1]$ around barycenter $G_i$ to scale down the projection area.


\begin{figure*}[htb]
    \centering
    \includegraphics[width=0.9\linewidth]{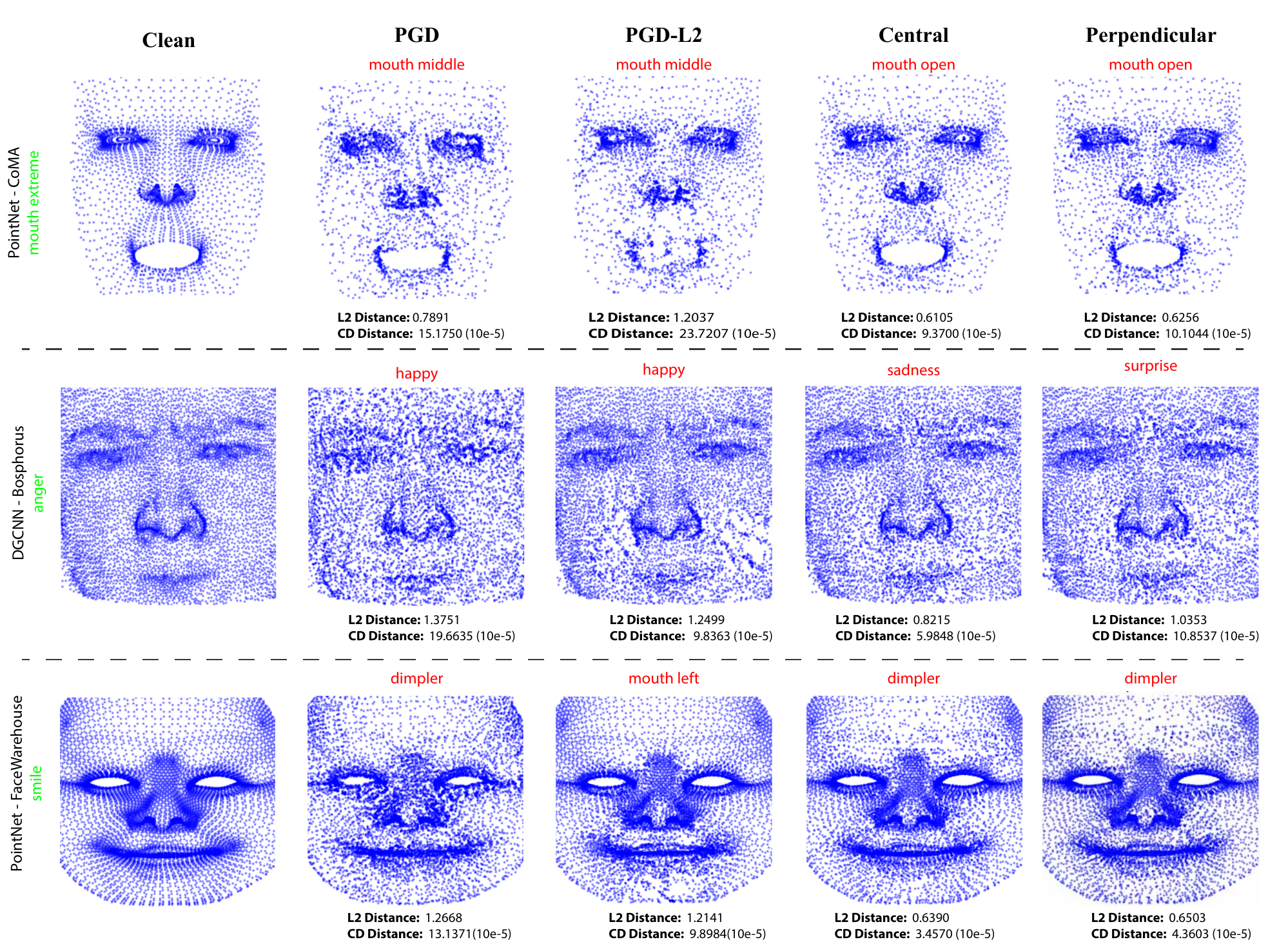}
    \caption{Example front-viewed point cloud images from  different datasets are given at each row.  Predictions obtained from  model denoted at left are written at the top of each image with green and red colors for correct and  incorrect predictions respectively. The columns represent a clean face, and its attacked versions by PGD, PGD-L2, $\epsilon$-mesh central projection, and $\epsilon$-mesh perpendicular projection from left to right. Also, distances between the attacked and clean point clouds are denoted below with each sample for $L_2$ and Chamfer distances.}
    \label{fig:tri-row results}
\end{figure*}

\section{EVALUATION}

\textbf{Datasets}. We have conducted our experiments on three well-known facial expression datasets: CoMA \cite{COMA:ECCV18}, Bosphorus \cite{savran2008bosphorus} and FaceWarehouse \cite{cao2013facewarehouse}. We have focused on 3D datasets where the mesh data is available and for the datasets like Bosphorus where the underlying mesh is missing, we have used Poisson surface reconstruction \cite{kazhdan2006poisson} as a simple triangular mesh estimator.

CoMA \cite{COMA:ECCV18} is a publicly available 4D facial expression mesh dataset. CoMA contains 12 facial expression (bare teeth, cheeks in, eyebrow, high smile, lips back, lips up, mouth down, mouth extreme, mouth middle, mouth open, mouth side, mouth up) sequences of 12 different subjects. Each subject performs facial expressions over a series of frames. For each sequence, we selected the peak frame and 4 more adjacent frames that are successive to the denoted frame. We randomly selected 10 subjects (600 meshes) for training and 2 subjects (120 meshes) for testing our models. 

Bosphorus 3D Database \cite{savran2008bosphorus} consists of various action units and emotions from 105 people. It contains both the RGB images and facial 3D coordinates for each pixel. The dataset has 6 emotion classes: anger, disgust, fear, happy, sad and surprise. We have split the dataset as 91 subjects for training and 14 subjects for the test set. 

FaceWarehouse dataset \cite{cao2013facewarehouse} contains reconstructed 3D mesh data of 150 individuals with 20 different facial poses (neutral, mouth stretch, smile, brow lower, brow raiser, anger, jaw left, jaw right, jaw forward, mouth left, mouth right, dimpler, chin raiser, lip puckerer, lip funneler, sadness, lip roll, grin, cheek blowing, eyes closed). We selected data from 126 individuals for training while preserving the other 24 for the test.

\textbf{Preprocessing}. We initially selected center of gravity for each triangle to form our point clouds from meshes. Then, a unit sphere scaling is applied. Subsequently, for the datasets containing the full head meshes, namely CoMA and FaceWarehouse, the back halves of the subjects' heads are deleted. After completing the preprocessing steps, it was observed that each sample from the Bosphorus, CoMA, and FaceWarehouse datasets consisted of approximately 12000, 4000, and 12000 point/mesh pairs, respectively. 


\textbf{Experimental Results}. We have compared our two suggested white box point cloud attack methods against Projected Gradient Descent method \cite{sun2021adversarially} with $L_{\infty}$ and $L_{2}$ metrics, namely PGD and PGD-L2 respectively.

\begin{table*}[btp]
\caption{Adversarial attack results on CoMA, Bosphorus and FaceWarehouse datasets}
\label{all-attack-results}
\begin{center}
\begin{tabular}{clccc||cc||cc||cc}
\multicolumn{5}{c||}{}                                                                                                                                                                                     & \multicolumn{2}{c||}{CoMA}                                                                                                                     & \multicolumn{2}{c||}{Bosphorus}                                                                                                                & \multicolumn{2}{c}{FaceWarehouse}                                                                                                             \\ \hline
\multicolumn{1}{c|}{Model}                     & \multicolumn{1}{l|}{Attack}                                                    & \multicolumn{1}{c|}{Eps}  & \multicolumn{1}{c|}{Alpha}  & Steps                & \multicolumn{1}{c|}{\begin{tabular}[c]{@{}c@{}}Clean \\ Acc (\%)\end{tabular}} & \begin{tabular}[c]{@{}c@{}}Attacked \\ Acc (\%)\end{tabular} & \multicolumn{1}{c|}{\begin{tabular}[c]{@{}c@{}}Clean \\ Acc (\%)\end{tabular}} & \begin{tabular}[c]{@{}c@{}}Attacked \\ Acc (\%)\end{tabular} & \multicolumn{1}{c|}{\begin{tabular}[c]{@{}c@{}}Clean \\ Acc (\%)\end{tabular}} & \begin{tabular}[c]{@{}c@{}}Attacked \\ Acc (\%)\end{tabular} \\ \hline \hline
\multicolumn{1}{c|}{\multirow{4}{*}{DGCNN\cite{wang2019dynamic}}}    & \multicolumn{1}{l|}{PGD}                                                       & \multicolumn{1}{c|}{0.01} & \multicolumn{1}{c|}{0.0004} & \multirow{4}{*}{250} & \multicolumn{1}{c|}{\multirow{4}{*}{79.17}}                                    & 0.0                                                          & \multicolumn{1}{c|}{\multirow{4}{*}{69.04}}                                    & 0.0                                                          & \multicolumn{1}{c|}{\multirow{4}{*}{98.96}}                                    & 0.0                                                          \\ \cline{2-4} \cline{7-7} \cline{9-9} \cline{11-11} 
\multicolumn{1}{c|}{}                          & \multicolumn{1}{l|}{PGD-L2}                                                    & \multicolumn{1}{c|}{1.25} & \multicolumn{1}{c|}{0.05}   &                      & \multicolumn{1}{c|}{}                                                          & 0.0                                                          & \multicolumn{1}{c|}{}                                                          & 0.0                                                          & \multicolumn{1}{c|}{}                                                          & 0.0                                                          \\ \cline{2-4} \cline{7-7} \cline{9-9} \cline{11-11} 
\multicolumn{1}{c|}{}                          & \multicolumn{1}{l|}{(\textbf{Ours}) $\epsilon$-mesh Central}                                       & \multicolumn{1}{c|}{1.00} & \multicolumn{1}{c|}{0.10}   &                      & \multicolumn{1}{c|}{}                                                          & 5.83                                                         & \multicolumn{1}{c|}{}                                                          & 3.57                                                         & \multicolumn{1}{c|}{}                                                          & 0.21                                                         \\ \cline{2-4} \cline{7-7} \cline{9-9} \cline{11-11} 
\multicolumn{1}{c|}{}                          & \multicolumn{1}{l|}{(\textbf{Ours}) $\epsilon$-mesh Perpendicular}                                 & \multicolumn{1}{c|}{1.00} & \multicolumn{1}{c|}{0.10}   &                      & \multicolumn{1}{c|}{}                                                          & 0.83                                                         & \multicolumn{1}{c|}{}                                                          & 0.0                                                          & \multicolumn{1}{c|}{}                                                          & 0.0                                                          \\ \hline \hline
\multicolumn{1}{c|}{\multirow{4}{*}{PointNet\cite{qi2017pointnet}}} & \multicolumn{1}{l|}{PGD}                                                       & \multicolumn{1}{c|}{0.01} & \multicolumn{1}{c|}{0.0004} & \multirow{4}{*}{250} & \multicolumn{1}{c|}{\multirow{4}{*}{71.67}}                                    & 0.0                                                          & \multicolumn{1}{c|}{\multirow{4}{*}{60.71}}                                    & 0.0                                                          & \multicolumn{1}{c|}{\multirow{4}{*}{88.96}}                                    & 0.0                                                          \\ \cline{2-4} \cline{7-7} \cline{9-9} \cline{11-11} 
\multicolumn{1}{c|}{}                          & \multicolumn{1}{l|}{PGD-L2}                                                    & \multicolumn{1}{c|}{1.25} & \multicolumn{1}{c|}{0.05}   &                      & \multicolumn{1}{c|}{}                                                          & 0.0                                                          & \multicolumn{1}{c|}{}                                                          & 0.0                                                          & \multicolumn{1}{c|}{}                                                          & 0.0                                                          \\ \cline{2-4} \cline{7-7} \cline{9-9} \cline{11-11} 
\multicolumn{1}{c|}{}                          & \multicolumn{1}{l|}{(\textbf{Ours}) $\epsilon$-mesh Central} & \multicolumn{1}{c|}{1.00} & \multicolumn{1}{c|}{0.10}   &                      & \multicolumn{1}{c|}{}                                                          & 0.83                                                         & \multicolumn{1}{c|}{}                                                          & 19.04                                                        & \multicolumn{1}{c|}{}                                                          & 14.38                                                        \\ \cline{2-4} \cline{7-7} \cline{9-9} \cline{11-11} 
\multicolumn{1}{c|}{}                          & \multicolumn{1}{l|}{(\textbf{Ours}) $\epsilon$-mesh Perpendicular}                                 & \multicolumn{1}{c|}{1.00} & \multicolumn{1}{c|}{0.10}   &                      & \multicolumn{1}{c|}{}                                                          & 0.0                                                          & \multicolumn{1}{c|}{}                                                          & 7.14                                                         & \multicolumn{1}{c|}{}   & 1.67                                                                 
\end{tabular}
\end{center}
\end{table*}

Table \ref{all-attack-results} illustrates the classification performances of DGCNN\cite{wang2019dynamic} and PointNet\cite{qi2017pointnet}   against PGD, PGD-L2, and our suggested attack methods. For all three aforementioned datasets, both PGD and PGD-L2 attacks were able to effectively degrade model performance to zero by disrupting the underlying structure, as their primary aim is not to preserve facial structural integrity. For our $\epsilon$-mesh attacks, perpendicular attack achieves less than 2\% accuracy in all cases except for prediction with PointNet in Bosphorus dataset. We assume that being due to errors in mesh estimation algorithm. On the other hand, $\epsilon$-mesh central attack achieves less than 20\% accuracy in all cases. Overall, our perpendicular attack performs better according to numerical results.

\begin{figure*}[ht]
    \centering
    \includegraphics[width=\linewidth]{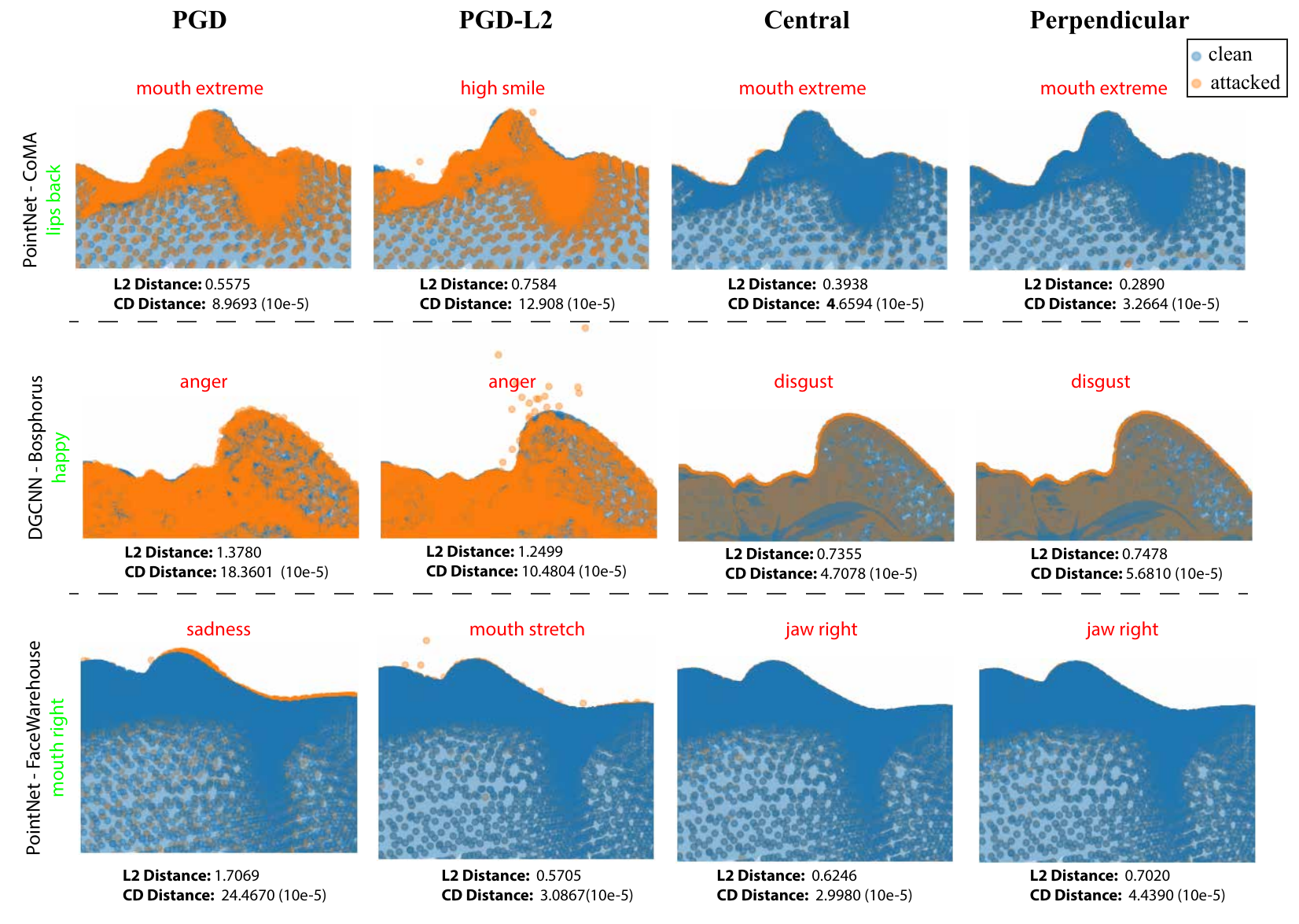}
    \caption{A side view of an example face mesh surface (blue triangles), clean point clouds (blue points), and attacked point clouds (orange points) for some samples from Coma, Bosphorus and FaceWarehouse datasets which are truly predicted by the given networks. Original network predictions (green texts) and attacked predictions (red texts) are also given. According to L2 and Chamfer distances, both $\epsilon$-Mesh Attack methods generally tend to corrupt the point cloud lesser than other methods.} 
    \label{fig:side-faces}
\end{figure*}

In the first row of Fig. \ref{fig:tri-row results}, we show that PGD attack heavily corrupts the data while PGD-L2 moves points over the empty regions such as the mouth. However, both of our methods keep the mouth region clear since it preserves the surface. For the same figure, as seen in the examples of second and third rows, PGD-L2 attack creates empty regions on the face surface during adversarial attack. For PGD examples on the same rows, outputs are rather noisy which is shown by high L2 and Chamfer distances to clean point clouds. Partial side views around the nose region is given in Figure \ref{fig:side-faces}. As seen in figure, the PGD attack pushes most of the points above surface while PGD-L2 shifts a few points to out of the surface furthermore. However, central and perpendicular $\epsilon$-mesh attacks are the only attacks that keep every point on the surface even after the adversarial perturbations.

\begin{figure*}[ht]
    \centering
    \includegraphics[width=\linewidth]{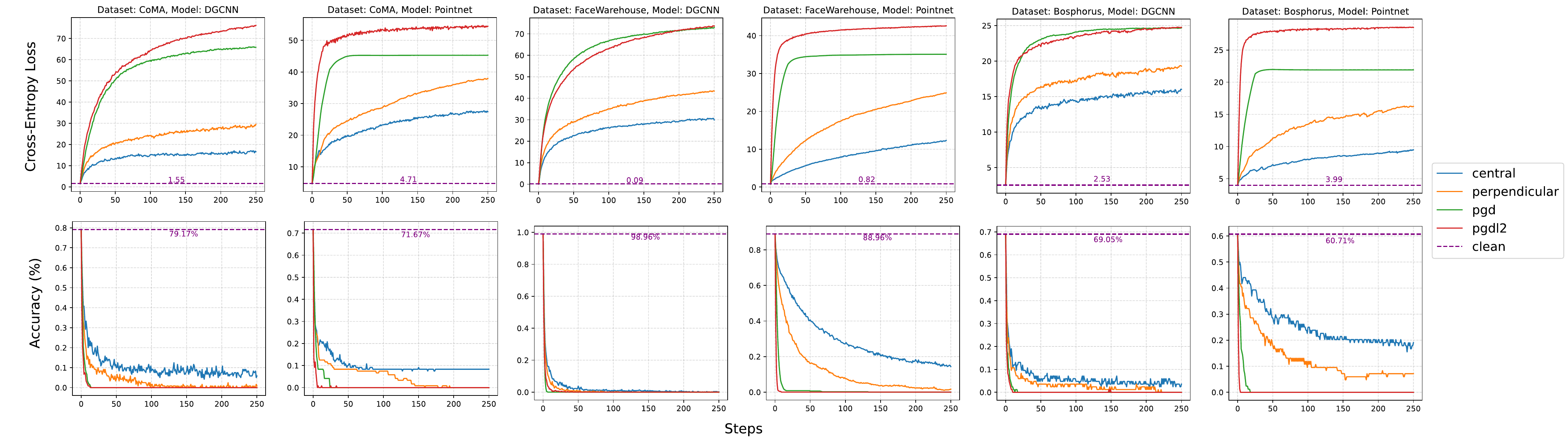}
    \caption{Adversarial attack loss results for varying number of steps. The left and right columns show results on DGCNN and PointNet respectively.}
    \label{fig:acc_loss_curves}
\end{figure*}

\textbf{Time complexity.} We have reported execution time for each model-dataset pair in Table \ref{time_table}. Our experiments show that the suggested two attack methods cost almost the same in terms of time, compared to other gradient based attacks like PGD. We apply a projection to each calculated gradient vector in each step. Thus, if we denote number of steps with $k$ and number of points with $n$, our time complexity would be $O(nk)$. Since CoMA dataset has less points than the other two, it costs less time to attack this dataset. Even though both PGD and $\epsilon$-mesh attack methods take equal time cost per step, the convergence rates differ as seen in Figure \ref{fig:acc_loss_curves}.

\begin{table}[htb]
\centering
\caption{Average execution time for adversarial attacks \\(250 steps)}
\label{all-attack-times}
\resizebox{0.45\textwidth}{!}{%
\begin{tabular}{c|ccc|}
\hline
\multicolumn{1}{|l|}{\multirow{3}{*}{\backslashbox{Attack}{Model\&\\Dataset}}} & \multicolumn{3}{c|}{\multirow{2}{*}{PointNet}}                 \\
\multicolumn{1}{|l|}{}                              & \multicolumn{3}{c|}{}                                          \\ \cline{2-4} 
\multicolumn{1}{|l|}{} & \multicolumn{1}{l|}{FaceWarehouse} & \multicolumn{1}{l|}{CoMA} & \multicolumn{1}{l|}{Bosphorus} \\ \hline
\multicolumn{1}{|c|}{Perpendicular}                 & \multicolumn{1}{c|}{1.80}  & \multicolumn{1}{c|}{1.37} & 1.75  \\ \hline
\multicolumn{1}{|c|}{Central}                       & \multicolumn{1}{c|}{1.95}  & \multicolumn{1}{c|}{1.47} & 1.90  \\ \hline
\multicolumn{1}{|c|}{PGD}                           & \multicolumn{1}{c|}{1.71}  & \multicolumn{1}{c|}{1.13} & 1.63  \\ \hline
\multicolumn{1}{|c|}{PGDL2}                         & \multicolumn{1}{c|}{1.70}  & \multicolumn{1}{c|}{1.12} & 1.67  \\ \hline
\multicolumn{1}{l|}{\multirow{3}{*}{}}              & \multicolumn{3}{c|}{\multirow{2}{*}{DGCNN}}                    \\
\multicolumn{1}{l|}{}                               & \multicolumn{3}{c|}{}                                          \\ \cline{2-4} 
\multicolumn{1}{l|}{}  & \multicolumn{1}{l|}{FaceWarehouse} & \multicolumn{1}{l|}{CoMA} & \multicolumn{1}{l|}{Bosphorus} \\ \hline
\multicolumn{1}{|c|}{Perpendicular}                 & \multicolumn{1}{c|}{25.17} & \multicolumn{1}{c|}{5.85} & 24.35 \\ \hline
\multicolumn{1}{|c|}{Central}                       & \multicolumn{1}{c|}{25.52} & \multicolumn{1}{c|}{5.98} & 24.59 \\ \hline
\multicolumn{1}{|c|}{PGD}                           & \multicolumn{1}{c|}{24.96} & \multicolumn{1}{c|}{5.74} & 24.31 \\ \hline
\multicolumn{1}{|c|}{PGDL2}                         & \multicolumn{1}{c|}{25.15} & \multicolumn{1}{c|}{5.73} & 24.24 \\ \hline
\end{tabular}%
}
\label{time_table}
\end{table}

\begin{figure}[ht]
    \centering
    \includegraphics[width=\linewidth]{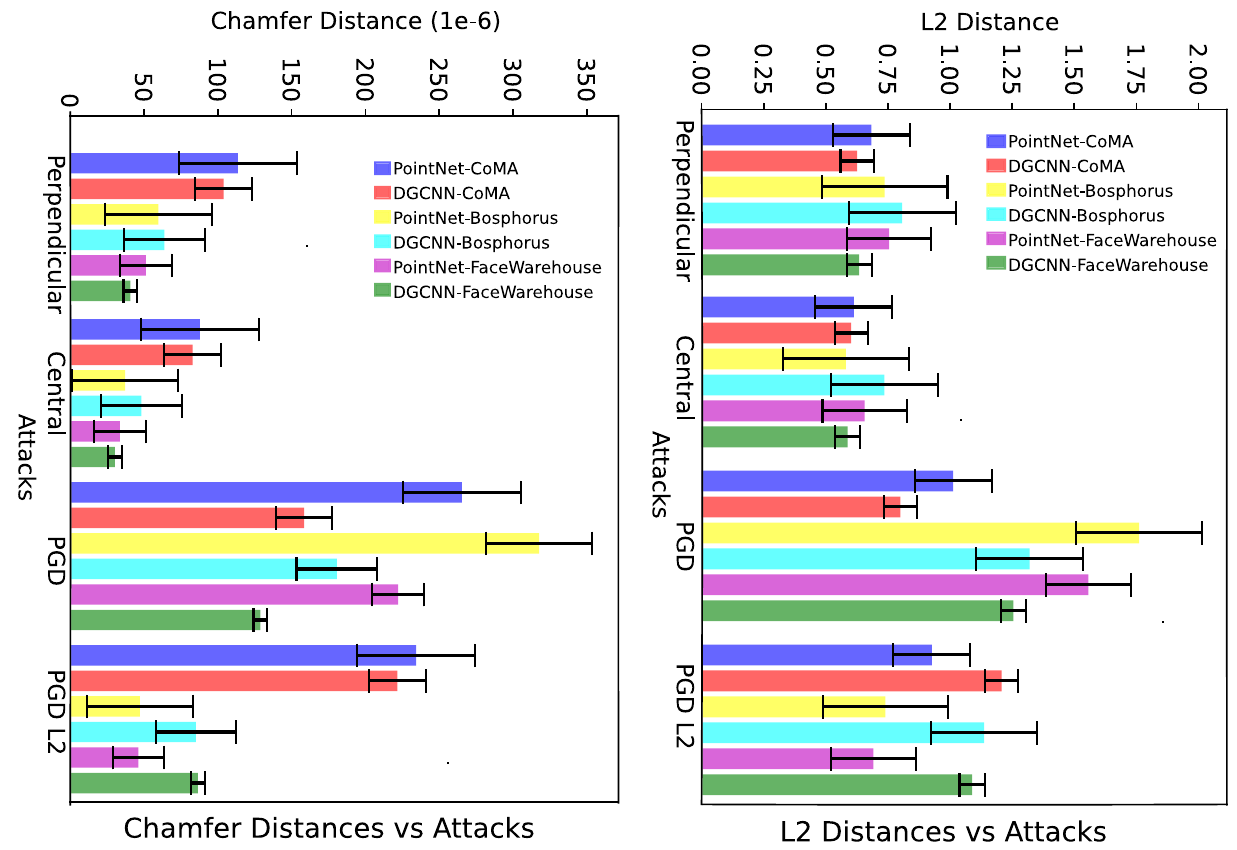}
    \caption{Adversarial attack results for varying number of steps (left column) and epsilon values (right column). The top and bottom rows show results on PointNet and DGCNN respectively.}
    \label{fig:barplots}
\end{figure}

\textbf{Perturbation distance.} We evaluated distances between the attacked point clouds and the clean point clouds using $L_2$ and Chamfer distances. Normally, it is not possible to calculate the $L_2$ distance between two point sets due to the unordered nature of the point clouds. However, we already have the correct point correspondes and perturbation vectors to calculate the $L_2$ distance. In Figure \ref{fig:barplots}, we have reported the distances for 6 different model-dataset pairs using bar notation to show means and standard deviations. For $L_2$ metric, our suggested perpendicular and central $\epsilon$-mesh attacks have a distance of 0.71 and 0.63 respectively, while PGD and PGD-L2 attacks have 1.28 and 0.97. For Chamfer distance, results are as following: 71.53 for perpendicular, 53.36 for central, 212.22 for PGD, 120.21 for PGD-L2. Allover, it is clear that our suggested methods perturbs the points at least 1.5 times less than the PGD based methods.

\begin{figure}[ht]
    \centering
    \includegraphics[width=\linewidth]{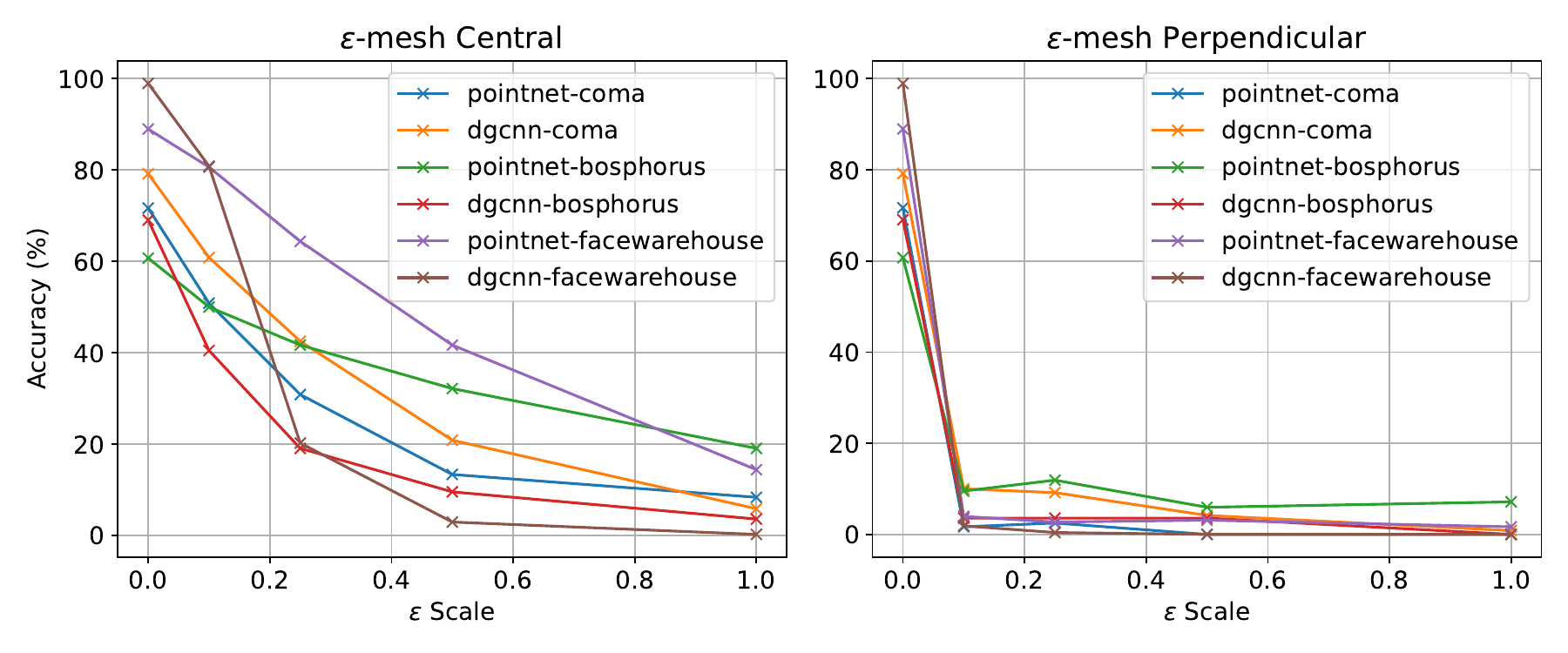}
    \caption{Adversarial attack accuracy results for different epsilon scales. The left and right graphs show results on $\epsilon$-mesh Central and Perpendicular attacks respectively.}
    \label{fig:ablation_acc_eps}
\end{figure}

\textbf{Ablation Study}. We test out our $\epsilon$ parameter which scales down the triangles to limit the perturbation into the center. In Fig. \ref{fig:ablation_acc_eps}, we reported results for $\epsilon \in \{0.1, 0.25, 0.5, 1.0\}$ while keeping the number of steps as 250 to show that scaling up the mesh boundaries increases the performance of attack exponentially. It is also clear that perpendicular projection performs well even under low values of $\epsilon$ such as $0.1$. 

Also in Fig. \ref{fig:acc_loss_curves}, we showed our experimental results for various number of steps in all attack methods. As iteration count increases, all methods perform better since we have an iterative optimization setup. PGD and PGD-L2 methods converge to 0\% accuracy just after 5 steps as they have softer perturbation boundaries as it can be deduced from the corruptions in Fig \ref{fig:side-faces}.

\section{DISCUSSION} \label{CONCLUSIONS AND FUTURE WORKS}

The primary aim of our paper was to introduce the $\epsilon$-mesh Attack, a new and distinct method in the field of adversarial attacks on 3D facial expression recognition models using point clouds and mesh data. The novelty of our approach lies in its ability to manipulate the points on their original mesh surfaces subtly while preserving its structural integrity, a feature not commonly emphasized in existing literature. To the best our knowledge, $\epsilon$-mesh attack is the first attack mechanism which considers the given mesh data alongside the point cloud. Therefore, the primary objective of our research was not to devise an attack method that surpasses others in terms of accuracy. Instead, our emphasis was on preserving the underlying mesh structure in the given point cloud. This focus is particularly significant as existing methods do not ensure that adversarial points conform to the surface, a critical aspect for realistic applications.

Our experimental results showed that the $\epsilon$-mesh Attack is capable of significantly reducing the performance of sophisticated deep learning models like DGCNN and PointNet. While our method does not achieve the aggressive performance degradation seen in methods like PGD, it offers a unique advantage in its subtlety and surface preservation. This finding is significant in highlighting the trade-off between aggressive attack strategies and the need for realistic, undetectable alterations in certain applications.

Preserving the surface structure of 3D faces in the context of adversarial attacks is crucial for several key reasons. In real-world applications, especially in fields like authentication, it is essential that any modifications to point clouds remain undetectable to the human eye. Preserving the surface structure ensures that the adversarial manipulations are subtle and do not alter the realistic appearance of the face, thus maintaining a high degree of stealthiness. In safety-critical areas like human-robot interaction it is important to obtain a robust prediction for the interactor's expression.

There are some potential countermeasures for defending against our proposed attack and other methods. Most straightforward one is using our method for an adversarial training procedure to obtain robust classification models. Other methods involve using sophisticated models for defensive strategies. These methods can be input preprocessing based defense models like methods in \cite{rusu2008towards} and \cite{zhou2019dup} to sanitize adversarial examples before classification.

There are some limitations to our proposed methods. First of all, our algorithms require relatively more steps to converge in a local minimum due to additional constraint of mesh boundaries. This limits our methods to having less steps to converge in real-time applications which could cause a performance decrease in terms of accuracy. Similarly, when compared to other methods with small number of steps, our methods fall relatively short on accuracy performance. Secondly, our method requires a mesh data which represents the true surface. In cases where the true mesh is not available, it is possible to use an estimation of the surface. However, it is possible for the surface estimator to affect the performance of our attack as shown with Bosphorus dataset.

\section{CONCLUSIONS AND FUTURE WORKS}
In this study, we proposed a 3D adversarial attack that preserves the structure of 3D faces while leading expression recognition models into error. Our adversarial attack limits the points by projecting the adversarial perturbations onto the triangles and reducing the 3D optimization domain into 2D triangles. This process is done by using one of two projection methods: Central and perpendicular projections. Central projection method leverages the positions of the triangles' center of mass to find the projection to an edge of the triangle. Perpendicular projection method projects the adversarial perturbations onto the closest point of triangle. We trained PointNet and DGCNN models for 3D facial expression recognition using the CoMA, Bosphorus, FaceWarehouse datasets and evaluated the robustness of these models with our attack. Results show our methods achieve protection of surface structure and point density while approaching the performance of 3d-ball attacks.  

As future work, our proposed two 3D adversarial point sampling methods can be leveraged over an adversarial setup to train robust classifiers. We would also like to span our 3D methods onto time domain to adjust into a 4D setup, which is natural domain of the experimented COMA \cite{COMA:ECCV18} dataset. We plan to address these topics in our future studies.



\clearpage

{\small
\bibliographystyle{ieee}
\bibliography{egbib}

\begin{thebibliography}{10}\itemsep=-1pt

\bibitem{afshar2016facial}
S.~Afshar and A.~Ali~Salah.
\newblock Facial expression recognition in the wild using improved dense trajectories and fisher vector encoding.
\newblock In {\em Proceedings of the IEEE Conference on Computer Vision and Pattern Recognition Workshops}, pages 66--74, 2016.

\bibitem{behzad2021disentangling}
M.~Behzad, X.~Li, and G.~Zhao.
\newblock Disentangling 3d/4d facial affect recognition with faster multi-view transformer.
\newblock {\em IEEE Signal Processing Letters}, 28:1913--1917, 2021.

\bibitem{TMPEH:CVPR:2023}
T.~Bolkart, T.~Li, and M.~J. Black.
\newblock Instant multi-view head capture through learnable registration.
\newblock In {\em Conference on Computer Vision and Pattern Recognition (CVPR)}, pages 768--779, 2023.

\bibitem{cao2013facewarehouse}
C.~Cao, Y.~Weng, S.~Zhou, Y.~Tong, and K.~Zhou.
\newblock Facewarehouse: A 3d facial expression database for visual computing.
\newblock {\em IEEE Transactions on Visualization and Computer Graphics}, 20(3):413--425, 2013.

\bibitem{carlini2017towards}
N.~Carlini and D.~Wagner.
\newblock Towards evaluating the robustness of neural networks.
\newblock In {\em 2017 ieee symposium on security and privacy (sp)}, pages 39--57. Ieee, 2017.

\bibitem{colombo2011umb}
A.~Colombo, C.~Cusano, and R.~Schettini.
\newblock Umb-db: A database of partially occluded 3d faces.
\newblock In {\em 2011 IEEE international conference on computer vision workshops (ICCV workshops)}, pages 2113--2119. IEEE, 2011.

\bibitem{deng2018ppfnet}
H.~Deng, T.~Birdal, and S.~Ilic.
\newblock Ppfnet: Global context aware local features for robust 3d point matching.
\newblock In {\em Proceedings of the IEEE Conference on Computer Vision and Pattern Recognition}, pages 195--205, 2018.

\bibitem{dhall2015video}
A.~Dhall, O.~Ramana~Murthy, R.~Goecke, J.~Joshi, and T.~Gedeon.
\newblock Video and image based emotion recognition challenges in the wild: Emotiw 2015.
\newblock In {\em Proceedings of the 2015 ACM on international conference on multimodal interaction}, pages 423--426, 2015.

\bibitem{duh2016facial}
D.-J. Duh, J.-C. Huang, S.-Y. Chen, S.~Su, H.~Zhang, and S.~Li.
\newblock Facial expression recognition based on spatio-temporal interest points for depth sequences.
\newblock {\em The Imaging Science Journal}, 64(7):396--407, 2016.

\bibitem{ekman1999basic}
P.~Ekman et~al.
\newblock Basic emotions.
\newblock {\em Handbook of cognition and emotion}, 98(45-60):16, 1999.

\bibitem{goodfellow2014explaining}
I.~J. Goodfellow, J.~Shlens, and C.~Szegedy.
\newblock Explaining and harnessing adversarial examples.
\newblock {\em arXiv preprint arXiv:1412.6572}, 2014.

\bibitem{hezroni2021deepbbs}
I.~Hezroni, A.~Drory, R.~Giryes, and S.~Avidan.
\newblock Deepbbs: Deep best buddies for point cloud registration.
\newblock In {\em 2021 International Conference on 3D Vision (3DV)}, pages 342--351. IEEE, 2021.

\bibitem{huang2022shape}
Q.~Huang, X.~Dong, D.~Chen, H.~Zhou, W.~Zhang, and N.~Yu.
\newblock Shape-invariant 3d adversarial point clouds.
\newblock In {\em Proceedings of the IEEE/CVF Conference on Computer Vision and Pattern Recognition}, pages 15335--15344, 2022.

\bibitem{huang2020feature}
X.~Huang, G.~Mei, and J.~Zhang.
\newblock Feature-metric registration: A fast semi-supervised approach for robust point cloud registration without correspondences.
\newblock In {\em Proceedings of the IEEE/CVF Conference on Computer Vision and Pattern Recognition}, pages 11366--11374, 2020.

\bibitem{kazhdan2006poisson}
M.~Kazhdan, M.~Bolitho, and H.~Hoppe.
\newblock Poisson surface reconstruction.
\newblock In {\em Proceedings of the fourth Eurographics symposium on Geometry processing}, volume~7, page~0, 2006.

\bibitem{kim2020softflow}
H.~Kim, H.~Lee, W.~H. Kang, J.~Y. Lee, and N.~S. Kim.
\newblock Softflow: Probabilistic framework for normalizing flow on manifolds.
\newblock {\em Advances in Neural Information Processing Systems}, 33:16388--16397, 2020.

\bibitem{kim2021setvae}
J.~Kim, J.~Yoo, J.~Lee, and S.~Hong.
\newblock Setvae: Learning hierarchical composition for generative modeling of set-structured data.
\newblock In {\em Proceedings of the IEEE/CVF Conference on Computer Vision and Pattern Recognition}, pages 15059--15068, 2021.

\bibitem{kurakin2018adversarial}
A.~Kurakin, I.~J. Goodfellow, and S.~Bengio.
\newblock Adversarial examples in the physical world.
\newblock In {\em Artificial intelligence safety and security}, pages 99--112. Chapman and Hall/CRC, 2018.

\bibitem{li2020deep}
S.~Li and W.~Deng.
\newblock Deep facial expression recognition: A survey.
\newblock {\em IEEE transactions on affective computing}, 13(3):1195--1215, 2020.

\bibitem{liu2017adaptive}
X.~Liu, B.~Vijaya~Kumar, J.~You, and P.~Jia.
\newblock Adaptive deep metric learning for identity-aware facial expression recognition.
\newblock In {\em Proceedings of the IEEE conference on computer vision and pattern recognition workshops}, pages 20--29, 2017.

\bibitem{liu20234d}
Y.-J. Liu, B.~Wang, L.~Gao, J.~Zhao, R.~Yi, M.~Yu, Z.~Pan, and X.~Gu.
\newblock 4d facial analysis: A survey of datasets, algorithms and applications.
\newblock {\em Computers \& Graphics}, 115:423--445, 2023.

\bibitem{lucey2010extended}
P.~Lucey, J.~F. Cohn, T.~Kanade, J.~Saragih, Z.~Ambadar, and I.~Matthews.
\newblock The extended cohn-kanade dataset (ck+): A complete dataset for action unit and emotion-specified expression.
\newblock In {\em 2010 ieee computer society conference on computer vision and pattern recognition-workshops}, pages 94--101. IEEE, 2010.

\bibitem{madry2017towards}
A.~Madry, A.~Makelov, L.~Schmidt, D.~Tsipras, and A.~Vladu.
\newblock Towards deep learning models resistant to adversarial attacks.
\newblock {\em arXiv preprint arXiv:1706.06083}, 2017.

\bibitem{mo2023dit}
S.~Mo, E.~Xie, R.~Chu, L.~Yao, L.~Hong, M.~Nie{\ss}ner, and Z.~Li.
\newblock Dit-3d: Exploring plain diffusion transformers for 3d shape generation.
\newblock {\em arXiv preprint arXiv:2307.01831}, 2023.

\bibitem{qi2017pointnet}
C.~R. Qi, H.~Su, K.~Mo, and L.~J. Guibas.
\newblock Pointnet: Deep learning on point sets for 3d classification and segmentation.
\newblock In {\em Proceedings of the IEEE Conference on Computer Vision and Pattern Recognition}, pages 652--660, 2017.

\bibitem{qi2017pointnet++}
C.~R. Qi, L.~Yi, H.~Su, and L.~J. Guibas.
\newblock Pointnet++: Deep hierarchical feature learning on point sets in a metric space.
\newblock In {\em Advances in neural information processing systems}, pages 5099--5108, 2017.

\bibitem{COMA:ECCV18}
A.~Ranjan, T.~Bolkart, S.~Sanyal, and M.~J. Black.
\newblock Generating {3D} faces using convolutional mesh autoencoders.
\newblock In {\em European Conference on Computer Vision (ECCV)}, pages 725--741, 2018.

\bibitem{richter2012facial}
M.~Richter, T.~Gehrig, and H.~K. Ekenel.
\newblock Facial expression classification on web images.
\newblock In {\em Proceedings of the 21st International Conference on Pattern Recognition (ICPR2012)}, pages 3517--3520. IEEE, 2012.

\bibitem{rusu2008towards}
R.~B. Rusu, Z.~C. Marton, N.~Blodow, M.~Dolha, and M.~Beetz.
\newblock Towards 3d point cloud based object maps for household environments.
\newblock {\em Robotics and Autonomous Systems}, 56(11):927--941, 2008.

\bibitem{savran2008bosphorus}
A.~Savran, N.~Aly{\"u}z, H.~Dibeklio{\u{g}}lu, O.~{\c{C}}eliktutan, B.~G{\"o}kberk, B.~Sankur, and L.~Akarun.
\newblock Bosphorus database for 3d face analysis.
\newblock In {\em Biometrics and Identity Management: First European Workshop, BIOID 2008, Roskilde, Denmark, May 7-9, 2008. Revised Selected Papers 1}, pages 47--56. Springer, 2008.

\bibitem{sun2021adversarially}
J.~Sun, Y.~Cao, C.~B. Choy, Z.~Yu, A.~Anandkumar, Z.~M. Mao, and C.~Xiao.
\newblock Adversarially robust 3d point cloud recognition using self-supervisions.
\newblock {\em Advances in Neural Information Processing Systems}, 34:15498--15512, 2021.

\bibitem{szegedy2013intriguing}
C.~Szegedy, W.~Zaremba, I.~Sutskever, J.~Bruna, D.~Erhan, I.~Goodfellow, and R.~Fergus.
\newblock Intriguing properties of neural networks.
\newblock {\em arXiv preprint arXiv:1312.6199}, 2013.

\bibitem{wang2019deep}
Y.~Wang and J.~M. Solomon.
\newblock Deep closest point: Learning representations for point cloud registration.
\newblock In {\em Proceedings of the IEEE/CVF International Conference on Computer Vision}, pages 3523--3532, 2019.

\bibitem{wang2019dynamic}
Y.~Wang, Y.~Sun, Z.~Liu, S.~E. Sarma, M.~M. Bronstein, and J.~M. Solomon.
\newblock Dynamic graph cnn for learning on point clouds.
\newblock {\em ACM Transactions on Graphics (tog)}, 38(5):1--12, 2019.

\bibitem{xiang2019generating}
C.~Xiang, C.~R. Qi, and B.~Li.
\newblock Generating 3d adversarial point clouds.
\newblock In {\em Proceedings of the IEEE/CVF Conference on Computer Vision and Pattern Recognition}, pages 9136--9144, 2019.

\bibitem{yang2018facial}
H.~Yang, U.~Ciftci, and L.~Yin.
\newblock Facial expression recognition by de-expression residue learning.
\newblock In {\em Proceedings of the IEEE conference on computer vision and pattern recognition}, pages 2168--2177, 2018.

\bibitem{yang2020facescape}
H.~Yang, H.~Zhu, Y.~Wang, M.~Huang, Q.~Shen, R.~Yang, and X.~Cao.
\newblock Facescape: a large-scale high quality 3d face dataset and detailed riggable 3d face prediction.
\newblock In {\em Proceedings of the ieee/cvf conference on computer vision and pattern recognition}, pages 601--610, 2020.

\bibitem{yang2019adversarial}
J.~Yang, Q.~Zhang, R.~Fang, B.~Ni, J.~Liu, and Q.~Tian.
\newblock Adversarial attack and defense on point sets.
\newblock {\em arXiv preprint arXiv:1902.10899}, 2019.

\bibitem{yin20084d}
L.~Yin, X.~Chen, Y.~Sun, T.~Worm, and M.~Reale.
\newblock A high-resolution 3d dynamic facial expression database.
\newblock In {\em 2008 8th IEEE International Conference on Automatic Face \& Gesture Recognition}, pages 1--6, 2008.

\bibitem{yin20063d}
L.~Yin, X.~Wei, Y.~Sun, J.~Wang, and M.~J. Rosato.
\newblock A 3d facial expression database for facial behavior research.
\newblock In {\em 7th international conference on automatic face and gesture recognition (FGR06)}, pages 211--216. IEEE, 2006.

\bibitem{zhang20233d}
J.~Zhang, L.~Chen, B.~Liu, B.~Ouyang, Q.~Xie, J.~Zhu, W.~Li, and Y.~Meng.
\newblock 3d adversarial attacks beyond point cloud.
\newblock {\em Information Sciences}, 633:491--503, 2023.

\bibitem{zhang2014bp4d}
X.~Zhang, L.~Yin, J.~F. Cohn, S.~Canavan, M.~Reale, A.~Horowitz, P.~Liu, and J.~M. Girard.
\newblock Bp4d-spontaneous: a high-resolution spontaneous 3d dynamic facial expression database.
\newblock {\em Image and Vision Computing}, 32(10):692--706, 2014.

\bibitem{zhou2019dup}
H.~Zhou, K.~Chen, W.~Zhang, H.~Fang, W.~Zhou, and N.~Yu.
\newblock Dup-net: Denoiser and upsampler network for 3d adversarial point clouds defense.
\newblock In {\em Proceedings of the IEEE/CVF International Conference on Computer Vision}, pages 1961--1970, 2019.

\end{thebibliography}
}

\end{document}